\pdfoutput=1    % apparently this is needed at the beginning for arXiv to process it
\documentclass[letterpaper, 10 pt, conference]{ieeeconf}

\IEEEoverridecommandlockouts                              % This command is only needed if 
\usepackage{multicol}
\usepackage{float}
\usepackage{comment}
\usepackage{array}
\usepackage{graphicx}
\usepackage{booktabs}
\usepackage{multirow}
\usepackage[normalem]{ulem}
\usepackage{soul}
\usepackage[ruled]{algorithm2e}
\usepackage[table]{xcolor}
\usepackage{hhline}

\usepackage{amsmath}								
\usepackage{amssymb}
\usepackage{latexsym}
\usepackage{bm}
\usepackage{commath}
\usepackage{float}
\usepackage{units}

\usepackage{tikz}
\usetikzlibrary{arrows,shapes,trees,calc}
\usetikzlibrary{backgrounds}
\usepackage{pgfplots}
\pgfplotsset{compat=1.14}
\usepgfplotslibrary{polar}
\usepgfplotslibrary{patchplots}

\usepackage{tikz-3dplot} %requires 3dplot.sty to be in same directory, or in your LaTeX installation
\definecolor{lightyellow}{RGB}{255,236,132}
\definecolor{lightgreen}{RGB}{161,239,10}
\definecolor{darkgreen}{RGB}{61,124,68}
\definecolor{lightblue}{RGB}{72,131,219}
\definecolor{darkblue}{RGB}{39,63,186}
\definecolor{plgreen}{RGB}{27,158,119}
\definecolor{plorange}{RGB}{217,95,2}
\definecolor{plpurple}{RGB}{117,112,179}
\definecolor{plpink}{RGB}{231,41,138}

\newcounter{RamCount}
\newcounter{DavidCount}
\usepackage{color}  % For Highlighting

\newcommand{\mtx}[1]{\begin{bmatrix} #1 \end{bmatrix}}
\newcommand{\aln}[1]{\begin{align} #1 \end{align}}

\newcommand{\xv}{x}
\newcommand{\uv}{u}
\newcommand{\Fv}{F}
\newcommand{\Real}{\mathbb{R}}
\newcommand{\F}{\mathcal{F}}

\usepackage{marginnote}

\title{\LARGE \bf
Nonlinear System Identification of Soft Robot \\ Dynamics Using Koopman Operator Theory
}

\author{Daniel Bruder, %<-this % stops a space
        C. David Remy, %
        and Ram Vasudevan, \emph{Member, IEEE} %
\thanks{*This material is supported by the Toyota Research Institute, and is based upon work supported by the National Science Foundation Graduate Research Fellowship Program under Grant No. 1256260 DGE. Any opinions, findings, and conclusions or recommendations expressed in this material are those of the author(s) and do not necessarily reflect the views of the National Science Foundation.}% <-this % stops a space
\thanks{The authors are with the Mechanical Engineering Department at the 
        University of Michigan, Ann Arbor, MI 48109, USA
        \{\tt\small bruderd, cdremy, ramv\}@umich.edu}%
}

\begin{document}

\maketitle
\thispagestyle{empty}
\pagestyle{empty}

\begin{abstract}
%% Soft robots are difficult to model, but safe to observe
Soft robots are challenging to model due in large part to the nonlinear properties of soft materials.
Fortunately, this softness makes it possible to safely observe their behavior under random control inputs, making them amenable to large-scale data collection and system identification.
This paper implements and evaluates a system identification method based on Koopman operator theory
in which models of nonlinear dynamical systems are constructed via linear regression of observed data by exploiting the fact that
every nonlinear system has a linear representation in the infinite-dimensional space of real-valued functions called observables.
% which offers a way to represent a nonlinear system as a linear system in the infinite-dimensional space of real-valued functions called observables
% and enables models of nonlinear systems to be constructed via linear regression of observed data.
%%
% The approach does not suffer from the limited convergence guarantees of other nonlinear system identification methods, which typically consist of solving a nonlinear non-convex optimization problem. 
The approach does not suffer from some of the shortcomings of other nonlinear system identification methods, which typically require the manual tuning of training parameters and have limited convergence guarantees.
A dynamic model of a pneumatic soft robot arm is constructed via this method, and used to predict the behavior of the real system.
The total normalized-root-mean-square error (NRMSE) of its predictions % over twelve validation trials 
is lower than
that of several other identified models including a neural network, NLARX, nonlinear Hammerstein-Wiener, and linear state space model.
\end{abstract}

\IEEEpeerreviewmaketitle

% Input sections here
\section{Introduction}
\label{sec:intro}

%% Soft robots can do cool things because they are compliant, and need to be controlled
Soft robots incorporate non-rigid materials into their morphology to facilitate compliant interactions with the external world.
This compliance allows them to manipulate delicate objects, adapt to unstructured environments, and interact safely with coexisting humans 
% without the sophisticated sensing and control algorithms required by their rigid-bodied counterparts 
\cite{rus2015design, majidi2014soft, lipson2014challenges}.
Since their utility is derived from their compliance, control methods that preserve and/or exploit this property are desirable. 

%% Models are necessary for control
Accurate models facilitate better control performance.
When an accurate model is available, predictive controllers can be built by using the model to calculate a feedforward term, then adding a feedback term to account for minor model uncertainty and disturbances.
% stabilize around that desired point/account for model uncertainty/error (i.e. make up the difference).
If an accurate model is unavailable, feedback must be relied upon more heavily.
This poses several problems for soft robots.
First, feedback requires sensing, but the morphology of soft robots precludes the use of most conventional sensors.
Suitable alternatives are currently in development \cite{felt2018modeling, felt2017inductance, yang2013gauge, kim2011epidermal}, but are not yet readily available.
Second, relying heavily on feedback to compensate for an inaccurate model has been illustrated to reduce the compliance of soft robotic systems \cite{della2017controlling}.
That is, excessive feedback negates the desirable compliance of a soft robot by replacing its natural dynamics with those of a slower, stiffer system.
Therefore, accurate models are required to control soft robots in a manner that reduces dependence on feedback and simultaneously preserves compliance.
% Since feedback is reactionary rather than anticipatory it can be slow to track a set point and prone to unwanted oscillations.
% \David{Other arguments for models:  Controller development/tuning, design (even thought that is not an argument for a model that has been identified on hardware), observer (compute un-sensed states), load estimation, ...}

%% FIGURE: Overview Diagram
\begin{figure}[t]
    \centering
    \includegraphics[width=\linewidth]{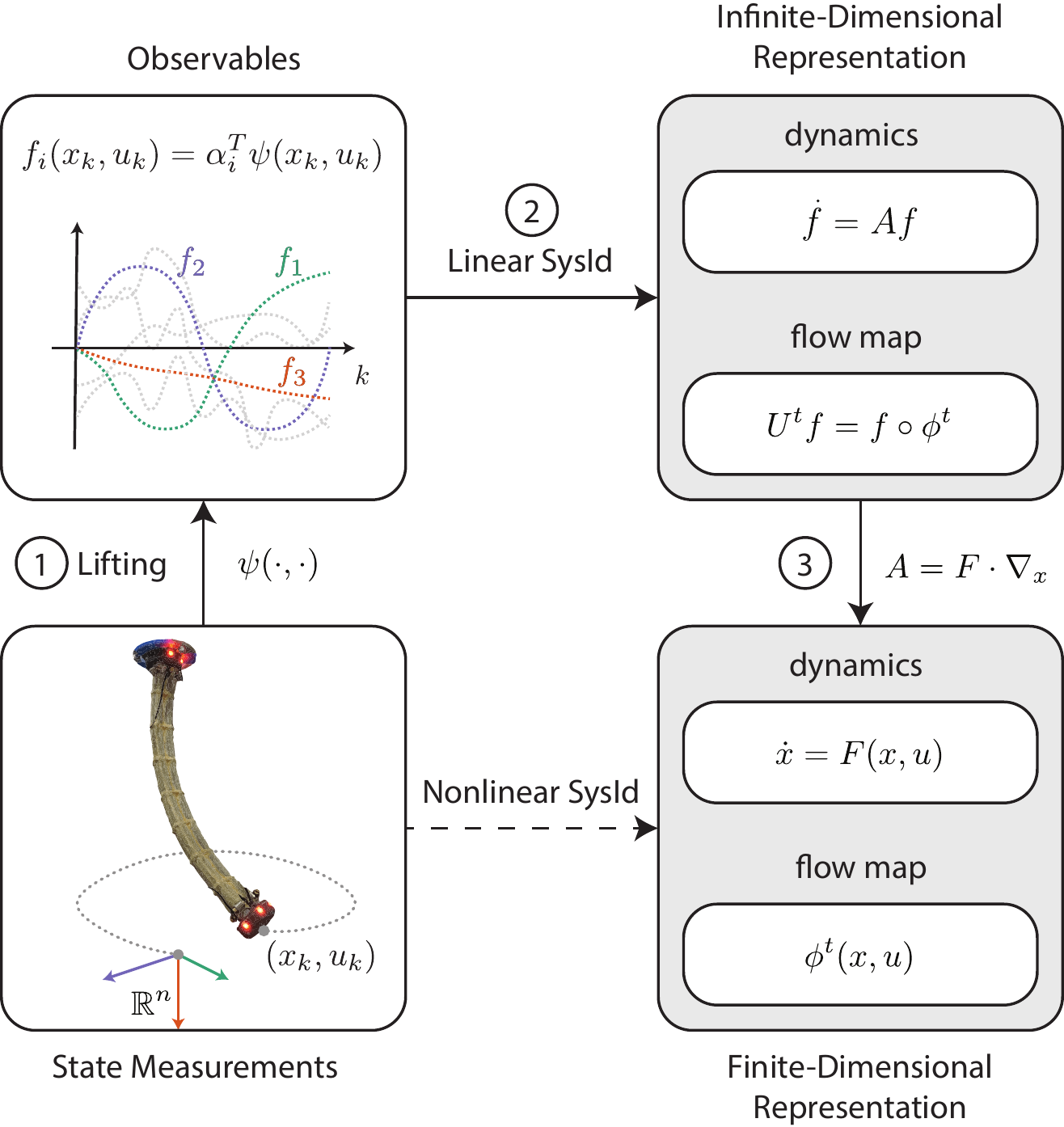}
    \caption{By providing an infinite-dimensional linear representation of a dynamical system, Koopman operator theory enables linear system identification of nonlinear systems. 
    This process proceeds in three steps, as described in Section \ref{sec:theory}:
    (1) Measured states of the system are lifted to the space of real-valued functions of the state and input.
    (2) Least-squares regression is performed on the lifted data to obtain an approximation of the Koopman operator, $U^t$.
    (3) An approximation of the nonlinear vector field $\Fv$ is obtained via its one-to-one relationship with the Koopman operator.}
    \label{fig:overview}
\end{figure}

% Physics-based and data-driven models
Models for soft robots can be separated into two categories: physics-based models and data-driven models.
Physics-based models are constructed from observations of component material properties and first-principles, while data-driven models are constructed from observations of system behavior.
Physics-based models have the ability to make predictions about a system's behavior before the system is constructed.
Thus, they are often used to inform the design and construction of soft robots intended for particular tasks \cite{bishop2015design, bruder2018iros, renda2014dynamic, felt2018closed, neppalli2009closed}.
%% Why physics-based models for soft robots are hard
However, the infinite degrees of freedom and nonlinear behavior of soft materials make it difficult to construct accurate physics-based models of soft robots without making simplifying assumptions such as constant curvature \cite{webster2010design, jones2006kinematics}, quasi-static \cite{george2018control, gravagne2003large, trivedi2008geometrically}, or simplified geometry \cite{bruder2017model, sedal2017constitutive, bishop2012parallel}.
These models only describe behavior well in the subset of robot configurations where the assumptions hold; hence, they are limited in applicability.
% This is bad for soft robots which are meant to be in contact with the world because they may often need to operate outside of regular specs
% For example, consider a linearization.
% By construction, the linear model describes system behavior well within some region of the equilibrium point
% \David{Talk to Audrey, she will have a good perspective about why it is hard to model soft robots.  I think other issues not mentioned here are computational cost, identification of material properties, influence of manufacturing tolerances,...}

%% Soft robots are particularly well suited for system identification
Data-driven models are more broadly applicable since they do not make structural assumptions about the system.
Data-driven models are constructed from observations of system behavior rather than from first-principles.
Hence, given sufficient data, they are capable of describing system behavior well over the entire range of observations.
% For the reasons stated above, it is typically not possible to construct globally accurate physics-based models of soft robots.
% In contrast, data-driven models are capable of describing system behavior well over the entire range of the observed data.
By virtue of their soft bodies, it is often possible to safely command arbitrary control inputs without risk of damaging a soft robot or nearby human operators.
Soft robots are therefore amenable to data collection over their entire operation range, making them particularly well suited for data-driven system identification techniques.

%% Nonlinear system identification is hard
To capture the characteristic nonlinear behaviors of most soft robots, identification of a nonlinear model is necessary.
Unfortunately, identifying a nonlinear model from data typically consists of solving a nonlinear, non-convex optimization problem, for which global convergence is not guaranteed \cite{boyd2004convex}.
Furthermore, most nonlinear system identification methods require the manual initialization and tuning of training parameters, which have an obscure impact on the resulting model.
A neural network, for example, may be able to capture the nonlinear behavior of a soft robot \cite{gillespie2018learning}; 
however, its accuracy depends on the number of hidden layers, number of nodes per layer, activation function, and termination condition used during training, which must be selected through trial and error until acceptable results are achieved.

%% Linear system identification is trivial
Linear model identification, on the other hand, does not suffer from the typical shortcomings of nonlinear identification 
since linear models can be identified via linear regression \cite{ljung1987system}.
However, linear models are poorly suited to capture the behavior of soft robots since their characteristic behavior is distinctly nonlinear \cite{rus2015design}.
% However, linear models have a limited ability to capture the nonlinear behavior exhibited by most soft robots \cite{rus2015design} \Dan{cite continuum paper}

% \Dan{deleted paragraph about neural network sysid paper}
%% Neural network has successfully been applied to soft robot
% Despite its challenges, nonlinear system identification of a soft robot has been performed in the past.
% In \cite{gillespie2018learning}, for instance, a Deep Neural Network (DNN) was used to model the behavior of a 1 DOF pneumatic soft robot arm.
% The resulting model predictive controller (MPC) with the neural network model acting as its predictor performed comparably to an MPC controller with a much more complicated physics-based model.
% While in this case the neural network model was effective, a weakness of neural network models in general is the extent to which the training process is heuristic in nature.
% A neural network model depends on the specific training parameters chosen (e.g. number of hidden layers, number of nodes per layer, activation function, termination condition), which must be selected through trial and error until acceptable results are achieved.

%% Koopman let's us do linear sysid (trivial) on a nonlinear system
In this work, we employ Koopman operator theory to generate nonlinear, data-driven models of soft robots via linear regression (see Fig. \ref{fig:overview}).
This approach relies on the idea of lifting nonlinear dynamical systems to an infinite dimensional space, where those systems have a linear representation.
In this space, it is possible to describe the dynamic behavior of a system by a linear operator rather than a nonlinear vector field \cite{budivsic2012applied}.
This linear operator, called the Koopman operator, is identified via linear regression \cite{williams2015data}, so it does not suffer from the convergence and tuning problems that are characteristic of neural networks and other nonlinear system identification methods.

%% Contributions (Why is this contribution significant?: Could enable real-time sysid/control, MPC)
Our primary contribution is demonstrating, on a real system, a data-driven method for constructing globally valid nonlinear models that does not require the manual tuning of multiple training parameters.
To do so, we apply the Koopman based system identification method described in \cite{mauroy2016linear} and \cite{mauroy2017koopman} to create a dynamic model of a soft robot arm and verify that it captures the system's true dynamic behavior better than the models generated by several other state-of-the-art nonlinear system identification methods including a neural network, a nonlinear auto-regressive with exogenous inputs model (NLARX), a nonlinear Hammerstein-Wiener model, and a state space model.
Koopman operator theory has previously been used to perform model-based control  of nonlinear systems \cite{Abraham-RSS-17}.
However, to the author's knowledge, this technique has never before been used to identify the dynamic model of a real system, much less a soft robot.
We believe that this system identification method applied to soft robots will enable the rapid development of new and effective control strategies by making accurate nonlinear dynamic models easier to construct.

%% Outline of the rest of this paper
The rest of this paper is organized as follows:
In Section \ref{sec:theory} we formally introduce the Koopman operator and describe the system identification method. 
In Section \ref{sec:experiment} we describe the soft robot and experimental procedure used for collecting input-output data of the system.
In Section \ref{sec:results} we summarize the results of applying various nonlinear system identification techniques to the collected data and compare the performances of the models generated. 
In Section \ref{sec:conclusion} concluding remarks and perspectives are provided.
\section{Koopman Operator Method for System Identification}
\label{sec:theory}

%% Overview of method
The system identification method utilized in this work exploits the fact that any finite-dimensional nonlinear system has an equivalent infinite-dimensional linear representation in the space of real-valued functions of the system's state and input.
In this space of real-valued functions, the (linear) Koopman operator describes the flow of functions along trajectories of the system.
The relationship between the finite and infinite dimensional representations of the system is bijective and well-defined \cite{lasota2013chaos}.
This enables us to approximate the Koopman operator via linear regression on observed data, then extract the equivalent nonlinear system representation.
% A thorough explanation of this method can be found in \cite{mauroy2016linear} or \cite{mauroy2017koopman}.
The remainder of this section summarizes the system identification method presented in \cite{mauroy2016linear} and \cite{mauroy2017koopman} applied to a system with known input, which is later employed and validated on a real soft robotic system.

%% Overview of lifting technique
\subsection{Overview of Lifting Technique}

%% System representation in state space
Consider a dynamical system
\begin{align}
    \dot{x} &= \Fv(\xv,\uv)    \label{eq:nlsys}
\end{align}
where $\xv \in \Real^n$ is the state of the system, $\uv \in \Real^m$ is the input and ${F}$ is continuously differentiable in $x$.
Denote by $\phi(t,\xv_0,\uv)$ the solution to \eqref{eq:nlsys} at time $t$ when beginning with the initial condition $\xv_0$ at time $0$ and a constant input $\uv$ applied for all time between $0$ and $t$.
For simplicity, we denote this map, which is referred to as the \emph{flow map}, by $\phi^t (\xv_0, \uv)$ instead of $\phi (t, \xv_0, \uv)$.

%% System representation in the space of observables
The system can be lifted to an infinite dimensional function space $\F = L^2(X \times U, \Real)$ where $X \subset \Real^n$ and $U \subset \Real^m$ are compact subsets and $ L^2(X \times U, \Real)$ is the space of square integrable real-valued functions with domain $X \times U$.
Elements of $\F$ are called \emph{observables}.
In $\F$, the flow of the system is characterized by the set %semigroup 
of Koopman operators 
$U^t : \F \to \F$, for each $t \geq 0$,
which describes the evolution of the observables $f \in \F$ along the trajectories of the system according to the following definition:
\begin{align}
    U^t f = f \circ \phi^t      
    % && \forall f \in \F, t \geq 0
    \label{eq:koopman}
\end{align}
As desired, $U^t$ is a linear operator even if the system \eqref{eq:nlsys} is nonlinear, since for $f_1, f_2 \in \F$ and $\lambda_1, \lambda_2 \in \Real$
\begin{align}
    \begin{split}
    U^t (\lambda_1 f_1 + \lambda_2 f_2) &= \lambda_1 f_1 \circ \phi^t + \lambda_2 f_2 \circ \phi^t \\
    &= \lambda_1 U^t f_1 + \lambda_2 U^t f_2.
    \end{split}
\end{align}
Thus the Koopman operator provides a linear representation of the flow of the system in the infinite-dimensional space of observables (see Fig. \ref{fig:overview}) \cite{budivsic2012applied}. 

%% Relationship between representations of the system
One can show that there is a one-to-one correspondence between the infinite-dimensional Koopman operator and finite-dimensional vector field.
To understand this relationship, consider the time derivative of an observable, $\dot{f}$, along trajectories of the system:
\begin{align}
    \dot{f}(x,u) &= \frac{\partial f}{\partial x} \frac{dx}{dt} + \frac{\partial f}{\partial u}\frac{du}{dt} \\
           &= \frac{\partial f}{\partial x} F(x,u) \\
           &= ( F \cdot \nabla_{\xv} ) f(x,u)
    \label{eq:fdot}
\end{align}
where the second relation follows since $u$ is held constant and where $\nabla_x$ is the gradient with respect to $x$.
Since this equation holds for all observables, we introduce the infinitesimal generator of the Koopman operator ${A:\F \to \F}$ \cite[Equation 7.6.5]{lasota2013chaos} which is defined in terms of the vector field $F$ as
\begin{align}
    % L &= \Fv \cdot \mtx{ \frac{\partial}{\partial \xv} & \frac{\partial}{\partial \uv} }^T
    A &= \Fv \cdot \nabla_{\xv}
    \label{eq:L2F}
\end{align}
The infinitesimal generator thus describes the dynamics of the observables along trajectories of the system (i.e. ${\dot{f}=Af}$). 
Recalling that the Koopman operator describes the flow of observables, one can show that the Koopman operater is associated with its infinitesimal generator via the following relation:
% where $\nabla_x$ is the gradient with respect to $x$, and $L$ is the infinitesimal generator of the Koopman operator  \cite[Section 7.6]{lasota2013chaos} \Ram{1. you never explain what the infinitestimal generator is...or even cite an appropriate reference. 2. it would help the reader if you described the domain and range of $L$} which satisfies
\begin{align}
    U^t &= e^{A t} = \sum_{k=0}^\infty \frac{t^k}{k!} A^k
    \label{eq:U2L}
\end{align}
In Section \ref{sec:step3} Equations \eqref{eq:L2F} \eqref{eq:U2L} are used to solve for the vector field $F$ when the Koopman operator $U^t$ is known.

%% Steps of sysid method
By providing a linear representation of a system with a one-to-one correspondence to its nonlinear representation, Koopman operator theory enables linear system identification of nonlinear systems. 
This process proceeds in three steps which are summarized by Algorithm \ref{alg:sysid}.
In step one, measured states of the system are lifted to the space of observables.
Step two consists of performing least-squares regression on the lifted data to obtain an approximation of the Koopman operator.
In step three, an approximation of the nonlinear vector field $\Fv$ is obtained using \eqref{eq:L2F} and \eqref{eq:U2L}.
The following three subsections describe each of these steps in more detail.

%% ALGORITHM
\begin{algorithm}
\SetAlgoLined
\KwIn{$ \{ (\xv_k,\uv_k),(\xv_{k+1},\uv_k) \} \text{ for } k = 1, ... , K$}
\textbf{Step 1:} Lift data via \eqref{eq:lift} \\
% \begin{algomathdisplay}
%     \psi(x_k,u_k) \hspace{15pt} \forall k \in 1,...,K
% \end{algomathdisplay} \\
\textbf{Step 2:} Approximate Koopman operator, $\bar{U}^{T_s}$ via \eqref{eq:Uapprox} \\
% \begin{algomathdisplay}
%     % \text{Step 2: (Approximate Koopman Operator) }
%     \bar{U}^{T_s} := \Psi_x^\dagger \Psi_y
% \end{algomathdisplay} \\
\textbf{Step 3:} Approximate Vector Field, $\bar{F}$ via \eqref{eq:logU} and \eqref{eq:Fbar} \\
% \begin{algomathdisplay}
%     \text{(a) } \bar{A} := \frac{1}{T_s} \log \bar{U}^{T_s} \hspace{20pt} \text{(b) } \bar{F} := \frac{\partial \Omega_{x}^\dagger}{\partial x} \underline{\bar{A}}^T \Omega_{x}
% \end{algomathdisplay} \\
\KwOut{$\bar{\Fv}$}
 \caption{Koopman-Based System Identification}
 \label{alg:sysid}
\end{algorithm}

%% STEP 1: Lifting the Data
\subsection{Step 1: Lifting the Data}   \label{sec:step1}

%% Overview of this step
The first step of the Koopman-based system identification method consists of converting empirical data into a form that can be used to identify a linear model in the space of observables.
Theoretically this process would consist of ``lifting'' state measurements into the infinite-dimensional space of observables $\F$.
To be implementable, however, measurements can only be lifted into a finite-dimensional subspace.
%% Approximation must be finite dimensional=
Define $\bar{\F} \subset \F$ to be the subspace of $\F$ spanned by $N$ linearly independent basis functions $\{ \psi_k \}_{k=1}^N$ 
(e.g. monomials, sinusoids, exponentials).
Any observable ${\bar{f} \in \bar{\F}}$ can be written as a linear combination of elements of the basis
\begin{align}
    \bar{f} &= \alpha_1 \psi_{1} + \cdots + \alpha_N \psi_N.
\end{align}
Note that the vector of coefficients ${\alpha = \mtx{\alpha_1 & \cdots & \alpha_N}^T}$ provides a \emph{vector representation} for $\bar{f} \in \bar{\F}$.
To evaluate $\bar{f}$ at a given state $\xv$ and constant input $\uv$, we introduce the \emph{lifting function} ${\psi}:\Real^n \times \Real^m \to \Real^N$ defined as:
\begin{align}
    {\psi}(\xv, \uv) = \mtx{ \psi_1(\xv, \uv) & \cdots & \psi_N(\xv, \uv)}^T
    \label{eq:lift}
\end{align}
Then, $\bar{f}(\xv, \uv)$ can be expressed in vector form as
\begin{align}
    \bar{f}(\xv, \uv) &= {\alpha}^T {\psi}(\xv, \uv).
    \label{eq:fxu}
\end{align}
We refer to ${\psi}(\xv, \uv)$ as an $N$-dimensional ``lifted'' version of ${(\xv, \uv)}$, since multiplying ${\psi}(\xv, \uv)$ by the vector representation of an observable yields the value of the observable at $(\xv, \uv)$.

%% STEP 2: Approximating the Koopman Operator
\subsection{Step 2: Approximating the Koopman Operator} \label{sec:step2}

The second step of the Koopman-based system identification method is to identify the Koopman operator that best describes the flow of the lifted versions of measured data points.
While the Koopman operator is theoretically infinite-dimensional, for practical purposes we identify a finite-dimensional approximation of it in $\bar{\F}$ which we denote by $\bar{U}^t$.
Note that $\bar{U}^t$ can be represented by an $N \times N$ matrix which operates on observables via matrix multiplication:
\begin{align}
    \bar{U}^t {\alpha} &= {\beta} 
    % && {\alpha}, {\beta} \in \F_N .
    \label{eq:Ubar}
\end{align}
where $\alpha, \beta$ are vector representations of observables in $\bar{\F}$.
Our goal is to find a $\bar{U}^t$ that describes the action of the infinite dimensional Koopman operator $U^t$ as accurately as possible in the $L^2$-norm sense on the finite dimensional subspace $\bar{\F}$  of all observables.
% We want to find $\bar{U}^t$ such that it describes the action of the infinite-dimensional Koopman operator $U^t$ as accurately as possible, i.e.
% \begin{align}
%     U^t f(\xv,\uv) &\approx ( \bar{U}^t {\alpha} )^T {\psi}(\xv, \uv)
% \end{align}
% \hl{for all $f \in \F_n$ with $\alpha$ as its vector representation}.
% From \eqref{eq:koopman}, the Koopman operator describes the flow of observables in $\F$.
Therefore, to perfectly mimic the action of $U^t$ acting on an observable in $\bar{\F} \subset \F$, the following should be true
\begin{align}
    ( \bar{U}^t {\alpha} )^T {\psi}(\xv, \uv) &=
    {\alpha}^T {\psi} \left( \phi^t(\xv,\uv), \uv \right).
    \label{eq:UbarEq}
\end{align}
Since this is a linear equation, it follows that for a given ${x \in \Real^n, \uv \in \Real^m}$, solving \eqref{eq:UbarEq} for $\bar{U}^t$ yields the best approximation of $U^t$ on $\bar{\F}$ in the $L^2$-norm sense:
\begin{align}
    \bar{U}^t = \left( {\psi}(\xv, \uv)^T \right)^\dagger {\psi}( \phi^t(\xv,\uv), \uv )^T
    \label{eq:Uapprox}
\end{align}
where superscript $\dagger$ denotes the least-squares pseudoinverse.

%% this whole paragraph should be removed...
% \sout{
% From \eqref{eq:koopman}, \eqref{eq:fxu}, and \eqref{eq:Ubar} we can approximate the effect of applying the infinite-dimensional Koopman operator $U^t$ to some $f \in \F_N$ using its finite-dimensional projection
% % \begin{align}
% %     U^t f(\xv,\uv) &\approx ( \bar{U}^t {\alphapeolpe} )^T {\psi}(\xv, \uv) =
% %     {\alpha}^T {\psi} \left( \phi^t(\xv,\uv), \uv \right) \\
% %     \label{eq:Utfapprox}
% % \end{align}
% where $\alpha$ is the vector representation of $f$.
% \Ram{you probably want to explain why each of these equations follow and I really hate when things are written down with approximations like that since with approximations without any explanation as to why its approximate.}
% From this relation the projection of the Koopman operator can be approximated
% % \begin{align}
% %     \bar{U}^t &\approx \left( {\psi}(\xv, \uv)^T \right)^\dagger {\psi}( \phi^t(\xv,\uv), \uv )^T
% %     \label{eq:Uapprox}
% % \end{align}
% where ${\psi}^\dagger$ denotes the pseudoinverse of ${\psi}$ \Ram{Same comment as earlier.}.
% }

%% How this is done on our system
To approximate the Koopman operator from a set of experimental data, we take $K+1$ discrete state measurements with sampling period $T_s$. Under the assumption that the control input is constant between samples, we separate the data into a set of $K$ so-called ``snapshot pairs'' of the form ${ \{(\xv_k,\uv_k),(y_{k},\uv_k)\} \in \Real^{(n \times m) \times 2} }$ where
\begin{align}
    y_{k} &= \phi^{T_s} (x_k, u_k) + \sigma_k
\end{align}
and $\sigma_k$ denotes measurement noise.
For our basis of $\bar{\F}$, we choose the basis of monomials of $\xv$ and $\uv$ with total degree less than or equal to $w$, which implies ${N=(n+m+w)!/\left((n+m)!w!\right)}$ \cite[Section III]{mauroy2016linear}. 
We then lift all of the snapshot pairs according to \eqref{eq:lift} and compile them into the following $K \times N$ matrices:
\begin{align}
    &\Psi_x = \mtx{ {\psi}(\xv_1, \uv_1)^T \\ \vdots \\  {\psi}(\xv_K, \uv_K)^T}
    &&\Psi_y = \mtx{ {\psi}(y_1, \uv_1)^T \\ \vdots \\  {\psi}(y_K, \uv_{K})^T}
\end{align}
 $\bar{U}^{T_s}$ is chosen so that it yields the least squares best fit to all of the observed data, which, following from \eqref{eq:Uapprox}, is given by 
\begin{align}
    \bar{U}^{T_s} &:= \Psi_x^\dagger \Psi_y.
\end{align}
% where ${\Psi}^\dagger$ denotes the least-squares pseudoinverse of ${\Psi}$.

%% STEP 3: Obtaining the vector field
\subsection{Step 3: Obtaining the Vector Field} \label{sec:step3}

The final step of the Koopman-based system identification method is to identify the nonlinear vector field by making use of the one-to-one correspondence between the infinite and finite dimensional system representations.
% Consider again the vector field $\Fv(\xv,\uv)=\mtx{F_1(\xv,\uv) & \cdots & F_n(\xv,\uv)}^T$.
% Each $F_i$ is a function of state and input that can be represented as a linear combination of the basis elements of $\F$.
As earlier, our goal is to find an $\bar{F}$ that describes the behavior of the vector field $F$ as accurately as possible in the $L^2$-norm sense on the finite dimensional subspace $\bar{\F}$.

% Our goal is to find a $\bar{U}^t$ that describes the action of the infinite dimensional Koopman operator $U^t$ as accurately as possible in the $L^2$-norm sense on the finite dimensional subspace $\F_N$  of all observables.

% As we did earlier, we restrict ourselves to represent $F$ using elements from the finite-dimensional subspace $\F_N$. 
% In particular, our goal is to find a finite dimensional representation 

% matrix of coefficients $C \in \Real^{n \times N}$ to 
% % \begin{align}
% %     C &= \mtx{ c_{1,1} & \cdots & c_{1,N} \\ \vdots & \ddots & \vdots \\ c_{n,1} & \cdots & c_{n,N} }
% %     \label{eq:C}
% % \end{align}

% the vector field resulting from this system identification method is only a linear combination of the basis elements of $\F_N$:
% \begin{align}
%     \Fv(\xv, \uv) &\approx C {\psi}(\xv, \uv)
%     \label{eq:F2C}
% \end{align}
% where $C$ is an $n \times N$ matrix of coefficients
% \begin{align}
%     C &= \mtx{ c_{1,1} & \cdots & c_{1,N} \\ \vdots & \ddots & \vdots \\ c_{n,1} & \cdots & c_{n,N} }
%     \label{eq:C}
% \end{align}
% Therefore, identifying the vector field reduces to the task of identifying $nN$ total coefficients $c_{i,j}$.

The vector field is related to the Koopman operator through its infinitesimal generator according to Equation \eqref{eq:L2F}.
% \Ram{sharp transition between the previous paragraph and the ensuing one}
With the approximation of the Koopman operator $\bar{U}^{T_s}$ found in Section \ref{sec:step2}, we can solve for the infinitesimal generator of the set of Koopman operators $\bar{A}$ by inverting \eqref{eq:U2L}:
\begin{align}
    \bar{A} &= \frac{1}{T_s} \log{ \bar{U}^t } \in \Real^{N \times N}
    \label{eq:logU}
\end{align}
where $\log$ denotes the principal matrix logarithm \cite[Chapter 11]{higham2008functions}.
Note that the principal matrix logarithm is only defined for matrices whose eigenvalues all have non-negative real components, and that $\bar{U}^t$ may have zero or negative eigenvalues when the number of data points is too small \cite{mauroy2016linear}.
Therefore this method might fail if the number of data points is insufficient.
In this instance, more system measurements can be taken to resolve the issue.

With $\bar{A}$ known, \eqref{eq:L2F} can be used to identify $\bar{F}$.
Consider $A$ applied to an observable $f \in \F$.
According to \eqref{eq:L2F}, this is equivalent to the inner product of the vector field $\Fv$ and the gradient of ${f}$ with respect to $x$:
\begin{align}
    A f(\xv,\uv) &= \frac{\partial f(\xv,\uv)}{\partial \xv} \Fv(\xv,\uv).
    \label{eq:L2Fxu}
\end{align}
% \Dan{i.e. the dynamics of the observable $\partial f / \partial t$}
Let $\alpha \in \Real^N$ be the vector representation $\bar{f}$, the projection of ${f}$ onto $\bar{\F}$.
Then from \eqref{eq:fxu} the finite-dimensional equivalent of \eqref{eq:L2Fxu} is given by
\begin{align}
    (\bar{A} {\alpha})^T {\psi}(\xv,\uv) &= {\alpha}^T \frac{\partial \psi(\xv, \uv)}{\partial \xv} \bar{F}.
    \label{eq:Lbar2C}
\end{align}
We seek the vector field $\bar{F}$ such that \eqref{eq:Lbar2C} holds as well as possible in the $L^2$-norm sense for all observed data.
Therefore we choose the least-square solution to \eqref{eq:Lbar2C} over the set of all observed data points $\left\{ (x_k,u_k) | k = 1,...,K \right\}$ which is given by
\begin{align}
    \bar{F} &= \mtx{ \frac{\partial \psi(\xv_1, \uv_1)}{\partial \xv} \\ \vdots \\ \frac{\partial \psi(\xv_K, \uv_K)}{\partial \xv} }^\dagger
    \mtx{ \bar{A}^T & \cdots & 0 \\ \vdots & \ddots & \vdots \\ 0 & \cdots & \bar{A}^T }
    \mtx{ \psi(x_1,u_1) \\ \vdots \\ \psi(x_K,u_K) }.
    \label{eq:Fbar}
\end{align}
% \begin{align}
%     \bar{F} &\approx \mtx{ \frac{\partial \psi(\xv_1, \uv_1)}{\partial \xv} \\ \vdots \\ \frac{\partial \psi(\xv_K, \uv_K)}{\partial \xv} }^\dagger
%         \mtx{ \bar{L}^T \\ \vdots \\ \bar{L}^T }.
% \end{align}
For a more thorough treatment of this process, see \cite{mauroy2016linear, mauroy2017koopman}.
\section{System Identification of a Soft Robot}
\label{sec:experiment}

To demonstrate and evaluate the performance of the method outlined in Section \ref{sec:theory}, we applied it to a continuously deformable soft robot arm and compared the resulting model to those generated by several other nonlinear system identification techniques.
In the following, we describe in detail the robotic hardware, experimental setup, measurement procedure, and the data processing involved in the system identification process, as well as the process by which performance was evaluated and compared across models.
% \David{In the following, we explain in detail the used robotic hardware, experimental setup, measurement proces, ...}This section describes the composition of the robot and recount the setup, measurement, and data pre-proccessing involved in the system identification process.

\subsection{Hardware Description}

\begin{figure}
    \centering
    \vspace{10pt}
    \includegraphics[width=0.8\linewidth]{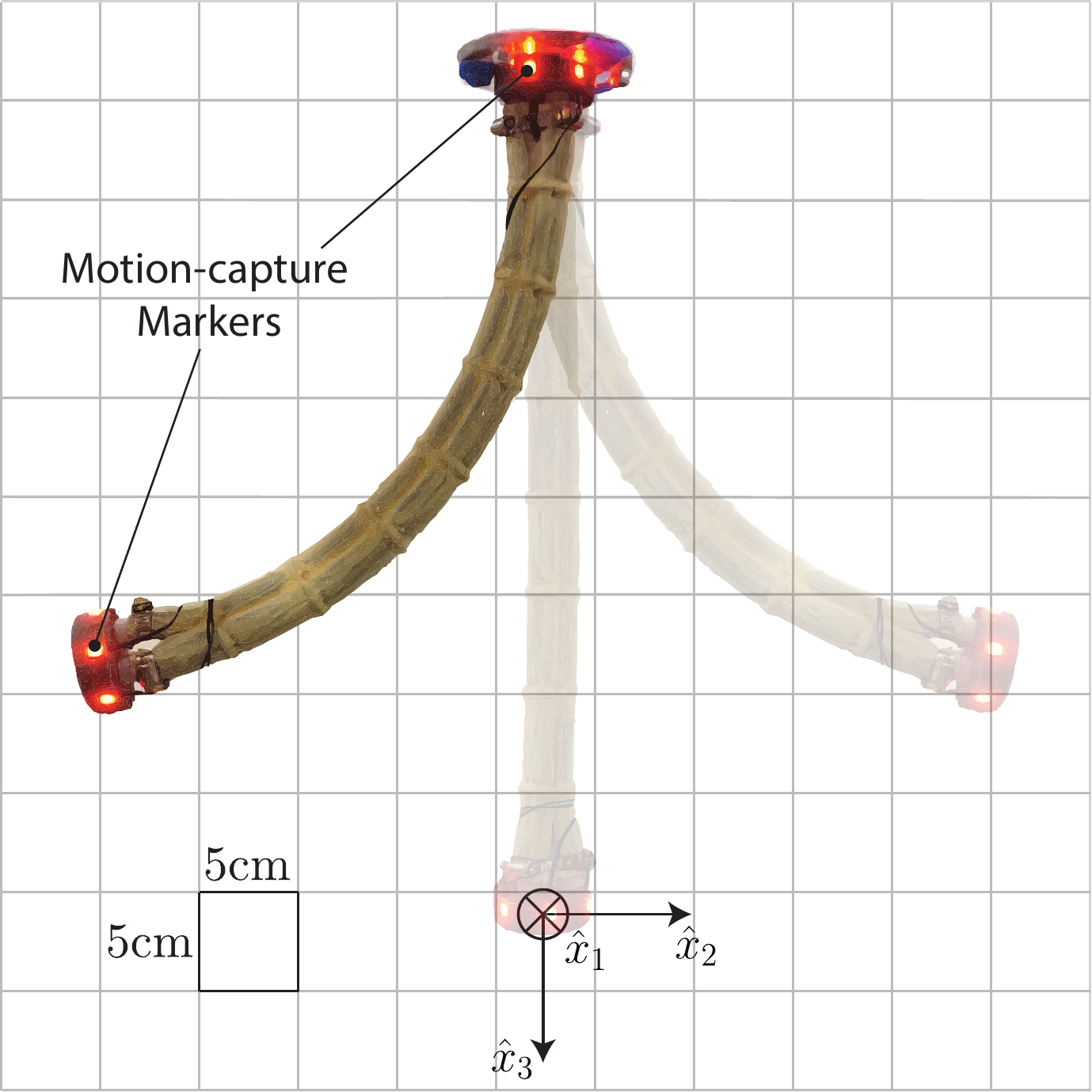}
    \caption{System identification was performed on a soft robot consisting of three PAMs adhered together. 
        Active motion capture markers on the base and end effector enabled tracking of the position and velocity of the end effector relative to the fixed global coordinate frame marked by unit vectors $\hat{x}_1,\hat{x}_2,\hat{x}_3$ where $\hat{x}_1$ is pointing into the page. The robot's range of motion given control inputs defined in \eqref{eq:input} is depicted.}
    \label{fig:flaccy}
\end{figure}

The soft robot used in this experiment consisted of three pneumatically driven McKibben actuators (also known as Pneumatic Artificial Muscles or PAMs) adhered together by latex rubber and connected to a common base mount on one end and to an end effector on the other (see Fig. \ref{fig:flaccy}).
During the trials, the pressure inside each actuator was varied using a pneumatic pressure regulator {(Enfield TR-010-g10-s)}, and the displacement and velocity of the end effector was measured at $60 \text{Hz}$ using a commercial motion capture system (Phase Space Impulse X2E).

%% state space representation of model
For this robot, it is our primary interest to control the motion of the end effector.
% \David{Try to use more declarative language.  We have discussed this in the past.  Other people do a lot of different things (and might disagree in your notion of primary interest), but you can simply say what your interest is: ``Our declared control goal for this setup is to dynamically control the kinematic motion of ...} With robotic manipulators, it is often of primary interest to dynamically control the motion of the end effector.
% \David{To this end, we chose...} 
Hence, we chose a state representation of the system which is capable of describing the dynamics of the end effector as an ordinary differential equation, namely the position and velocity of the end effector with respect to a global coordinate frame, as shown in Fig.~\ref{fig:flaccy}
% Hence, we chose to represent the state of our soft robot as the position and velocity of the end effector with respect to a global coordinate frame, as shown in Fig. \ref{fig:flaccy}
\begin{align}
    \vec{x} &= \mtx{ x_1 & x_2 & x_3 & \dot{x}_1 & \dot{x}_2 & \dot{x}_3 }^T
    \label{eq:state}
\end{align}
% It is worth noting that our chosen set of state variables may not be rich enough to uniquely describe the actual state of the system, e.g. there may be multiple body configurations associated with the same end effector location.
% In this case, the learned models will provide a best approximation of the dynamics of the chosen variables based on the training data, but won't satisfy the formal requirements of a state-space model.

\subsection{Randomized Control Input}

To generate a representative sampling of the system's behavior over its entire operation range, a randomized input was applied.
The control input into the system $\uv$ was a set of three $0-10 \text{V}$ signals into the pressure regulators corresponding to actuator pressures of $\approx 0-140 \text{ kPa}$
\aln{
    \uv (t) &= \mtx{ u_1 (t) & u_2 (t) & u_3 (t) }^T,   && u_i \in [0,10]
}
Before each trial, a $3 \times K_u$ table $\Upsilon$ of uniformly distributed random numbers between zero and ten was generated to be used as an input lookup table.
$K_u$ was chosen to be large enough to provide inputs over the entire length of each trial.
Each control input was smoothly varied between elements in consecutive columns of the table over a transition period $T_u$ with a time offset of $T_u / 3$ between each of the three control signals,
\aln{
    u_i (t) &= \frac{(\Upsilon_{i,k+1} - \Upsilon_{i,k})}{T_u} \left( t + \frac{(i-1) T_u}{3} \right) + \Upsilon_{i,k}
    \label{eq:input}
}
where $k = \text{floor}\left( {t} / {T_u} \right)$ is the current index into the lookup table at time $t$. 
% (see Fig. \ref{fig:randInput}).

%% Figure: Randomized Inputs
% \begin{figure}
%     \centering
%     \includegraphics[width=\linewidth]{figures/randInput3dim.pdf}
%     \caption{Caption: Randomized control inputs \David{I would add the dotted line at t+4/3T.}}
%     \label{fig:randInput}
% \end{figure}

\subsection{Data Collection}

% %% TABLE: Trial parameters
% \begin{table}[h]
%     \centering
%     \caption{Data Collection Parameters}
%     \begin{tabular}{|c||c|c|}
%         \hline
%          & $\bm{T_u}$ \textbf{(s)} & \textbf{Length (min)} \\
%         \hline
%          Trial 1 & 3 & 40:06 \\
%          Trial 2 & 3.5 & 31:26 \\
%          Trial 3 & 4 & 8:16 \\
%          Trial 4 & 5 & 26:09 \\
%          Trial 5 & 5 & 28:38 \\
%          Trial 6 & 8 & 31:35 \\
%         \hline
%     \end{tabular}
%     \label{tab:trialParams}
% \end{table}

%% TABLE: RMSE results table
\begin{table}
    \rowcolors{2}{white}{gray!25}
    \setlength\tabcolsep{5pt} % default value: 6pt
    \centering
    \caption{Data Collection Parameters}
    \begin{tabular}{|c|c|c|c|c|c|c|}
        \hline
        \rowcolor{white} 
        & \multicolumn{6}{c |}{\textbf{Trial}} \\
        \hhline{~------} \rowcolor{white}
        \multirow{-2}{*}{} & $1$ & $2$ & $3$ & $4$ & $5$ & $6$ \\
        \hline
        % RESULTS FOR ROBOT A
        \textbf{Length (min)}     &  40:06  &  31:26  &  8:16  &  26:09  &  28:38  &  31:35 \\
        \textbf{$\bm{T_u}$ (s)}  &  3.0  &  3.5  &  4  &  5  &  5  &  8 \\
        \hline
        % % RESULTS FOR ROBOT B
        % \cellcolor{white} & Koopman & & & & & & & & \\
        % \cellcolor{white} & Neural Net & & & & & & & & \\
        % \cellcolor{white} & State Space & & & & & & & & \\
        % \cellcolor{white} & Ham.-Weiner & & & & & & & & \\
        % \multirow{-5}{*}{\cellcolor{white} \rotatebox[origin=c]{90}{\textbf{Robot B}}}
        % & NLARX & & & & & & & & \\
        % \hline
    \end{tabular}
    \label{tab:trialParams}
\end{table}

%% TABLE: Trial parameters (ONLY IF TWO ROBOTS ARE USED)
% \begin{table}[h]
%     \centering
%     \caption{Data Collection Parameters}
%     \begin{tabular}{|c||c|c||c|c|}
%         \hline
%         & \multicolumn{2}{c|}{Robot A} & \multicolumn{2}{c|}{Robot B} \\
%         \hline
%          & $\bm{T_u}$ \textbf{(s)} & \textbf{Length (min)} & $\bm{T_u}$ \textbf{(s)} & \textbf{Length (min)} \\
%         \hline
%          Trial 1 & 3   & 40:06 & 2   & \\
%          Trial 2 & 3.5 & 31:26 & 3   & \\
%          Trial 3 & 4   & 8:16  & 3.5 & \\
%          Trial 4 & 5   & 26:09 & 4   & \\
%          Trial 5 & 5   & 28:38 & 5   & \\
%          Trial 6 & 8   & 31:35 & 6   & \\
%         \hline
%     \end{tabular}
%     \label{tab:trialParams2}
% \end{table}

Data collection proceeded in 6 trials each lasting an average of $\approx27$ minutes.
The input transition period $T_u$ varied from trial to trial, taking values between 3 and 8 seconds (see Table~\ref{tab:trialParams} for the specific values for each trial).
% The specific testing parameters for each trial can be found in Table \ref{tab:trialParams}.
After collecting data, raw position and velocity measurements were put through a moving average filter with window size of $1$s to reduce noise, then sampled uniformly with a period $T_s = 0.02$ seconds.
Sampled velocity measurements were put through a second moving average filter with window size of $1$ second due to the higher noise content of the velocity signal.
%% Partitioning of trail data into training and validation sets
The time-series data from each trial was partitioned into training and validation sets. 
Three 10 second validation sets were extracted from each trial and the remainder of the trial data was used for training.

%% FIGURE: Showing what flaccy looks like

%% Model Comparison
\subsection{Model Comparison}
\label{sec:modelcomparison}

%% Performance was evaluated by comparing simulations to real measurements
We generated a state space model from the technique described in section \ref{sec:theory} using a monomial basis of maximum degree $w=3$ and the collected training data.
We then evaluated its accuracy by comparing model simulations to each of the validation data sets (Fig.~\ref{fig:koopmanSim}).
Goodness of fit for the trajectory of a state $y$ was calculated using the normalized root-mean-square error (NRMSE), defined:
%% Normalized Root Mean-Square Error
\aln{
    \text{RMSE} &= \sqrt{ \frac{\sum_{k=1}^{N_\text{total}} \left( y_k - \hat{y}_k \right)^2}{N_\text{total}} } \\
    \text{NRMSE} &= \left( \frac{\text{RMSE}}{y_\text{max} - y_\text{min}} \right) \cdot 100 \%
    % 1 - \frac{ \| y - \hat{y}  \| }{ \| y - \text{mean}(y) \| }
}
where $\hat{y}$ is the simulated value of the state, $N_\text{total}$ is the total number of points, and $y_{\text{min/max}}$ are the measured minimum/maximum values of the state observed over all trials.

%% For comparison, other sysid methods were also used on the same set of data
The performance of our model was benchmarked against a linear state space, nonlinear Hammerstein-Wiener, nonlinear auto-regressive with exogenous inputs (NLARX), and a feedforward neural network model.
The models were trained and evaluated on the non-lifted time-series data from the experiments described in Section \ref{sec:experiment}, and generated using either the Matlab System Identification Toolbox or Neural Network Toolbox \cite{MATLAB:2017}.
The state space model was generated using the subspace method \cite[Chapter 7]{ljung1987system} and specified to be 6 dimensional, i.e. the same dimension as the state defined in \ref{eq:state}.
The neural network model was trained using the Levenberg-Marquardt backpropogation algorithm and sigmoid activation functions.
It was trained several times using combinations of 10-30 hidden neurons and 1-10 delays.
Only the results for the best of these models, corresponding to 10 hidden neurons and 10 delays, is displayed in Fig. \ref{fig:comparison} and Table \ref{tab:RMSE}.
\section{Results}
\label{sec:results}

\begin{figure}
    \centering
    \includegraphics[width=0.9\linewidth]{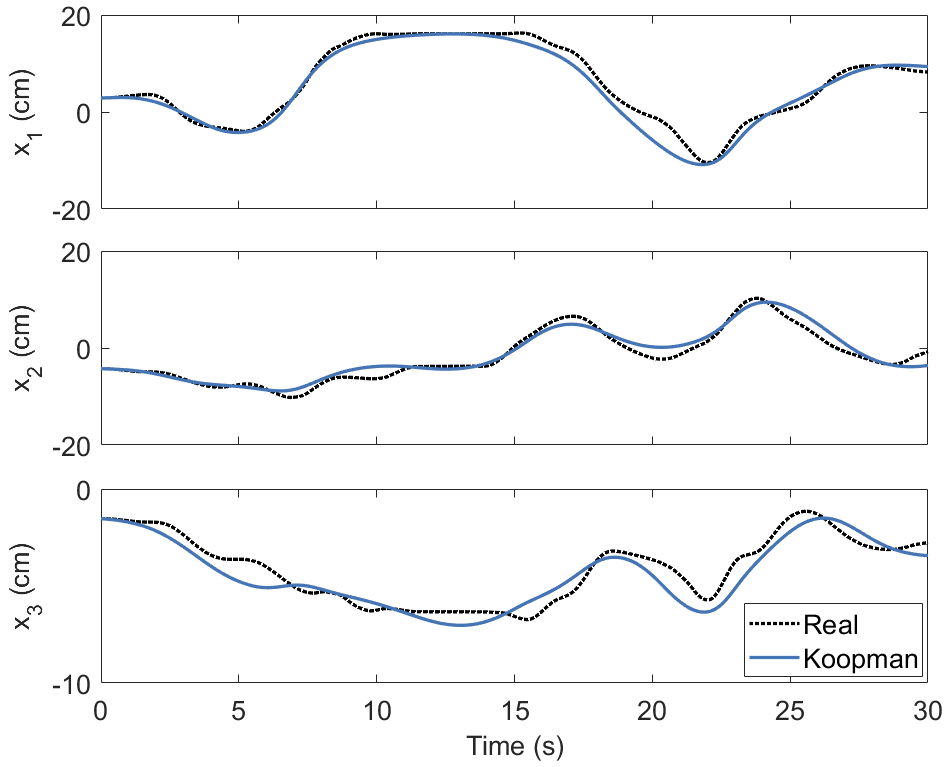}
    \caption{
    The measured position of the robot end effector over a 30 second time window (black,dotted) superimposed with the position predicted by the Koopman-based model (blue) given the same initial condition and control inputs. Coordinates are defined with respect to the global coordinate frame depicted in Fig. \ref{fig:flaccy}.}
    \label{fig:koopmanSim}
\end{figure}

\begin{figure}
    \centering
    \includegraphics[width=\linewidth]{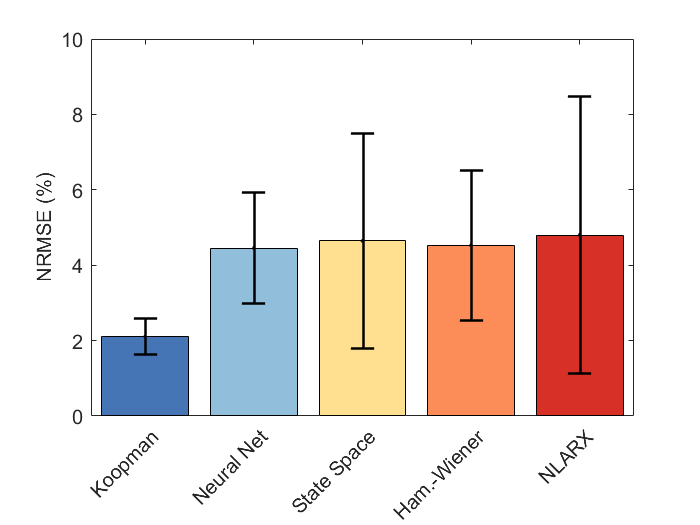}
    \caption{Shown is the total NRMSE averaged across all states for each of the models, with the standard deviation designated by the black bar. The average NRMSE of the Koopman-based model is less than half of that of the other models, with a standard deviation of less than one third of the other models.}
    \label{fig:comparison}
\end{figure}

The model generated by the Koopman system identification method has a total RMSE averaged across position states and velocity states of {5.98~mm} and {3.66~mm/s}, respectively.
As shown in Table \ref{tab:RMSE}, this corresponds to a total NRMSE averaged across all states of 2.1\%. 
By this metric, it performs more than twice as well as the best competing linear and nonlinear models which have average NRMSEs of 4.6\% and 4.5\%, respectively (see Fig.~\ref{fig:comparison}).
% Fig. \ref{fig:comparison} shows the NRMSE of each model over all states and all trials as compared to the validation data.
The Koopman model also exhibits the smallest standard deviation of the NRMSE across states.
This implies that the Koopman model more consistently captures the real behavior of all six states of the system, rather than just a subset of them.
Fig. \ref{fig:koopmanSim} illustrates the ability of the Koopman-based model to predict the position of the end effector over a 30 second time horizon.

% \David{I feel your results should start with what is obvious to you: IT WORKS. It converges.  You get a result. The result can predict behavior with an accuracy of XYZ.  It took x long to get this result.  If you used different seeds, you always get the same answer (because linear fitting).  Don't forget that you are doing this for the first time and it is important to note that it works.  People will ask themselves how well it works and how they can use it.  Do you need to tune stuff?  Do you need to carefully select parameters or initial guesses?}
% \David{Can you quantify the level of `non-linearity' of your robot?  This would be very valuable for a reader who wants to understand if this nonlinear approach is even necessary.  Can you discuss which of the other methods are linear/ nonlinear.}

%% FIGURE: Koopman simulation plot

%% FIGURE: Comparison bar graph

%% TABLE: RMSE results table
\begin{table}
    \rowcolors{2}{white}{gray!25}
    \setlength\tabcolsep{5pt} % default value: 6pt
    \centering
    \caption{Total NRMSE (\%) over all validation trials}
    \begin{tabular}{|c|c|c|c|c|c|c|c|c|}
        \hline
        \rowcolor{white} 
        & \multicolumn{6}{c |}{\textbf{States}} & & \textbf{Std.} \\
        \hhline{~------~~} \rowcolor{white}
        \multirow{-2}{*}{\textbf{Model}} & $x_1$ & $x_2$ & $x_3$ & $x_4$ & $x_5$ & $x_6$ & \multirow{-2}{*}{\textbf{Avg.}} & \textbf{Dev.} \\
        \hline
        % RESULTS FOR ROBOT A
        Koopman     &  2.4  &  2.0  &  2.9  &  1.7  &  1.5  &  2.0 & 2.1 & 0.5 \\
        Neural Net  &  5.8  &  4.0  &  6.6  &  3.9  &  2.8  &  3.5 & 4.5 & 1.5 \\
        State Space &  5.1  &  3.1  &  9.9  &  3.0  &  1.8  &  4.8 & 4.6 & 2.9 \\
        Ham.-Weiner &  7.0  &  4.5  &  6.9  &  3.0  &  2.3  &  3.1 & 4.5 & 2.0 \\
        % \multirow{-5}{*}{\cellcolor{white} \rotatebox[origin=c]{90}{\textbf{Robot A}}}
        NLARX       &  5.0  &  3.0 &  12.0  &  3.8  &  2.1  &  2.8 & 4.8 & 3.7 \\
        \hline
        % % RESULTS FOR ROBOT B
        % \cellcolor{white} & Koopman & & & & & & & & \\
        % \cellcolor{white} & Neural Net & & & & & & & & \\
        % \cellcolor{white} & State Space & & & & & & & & \\
        % \cellcolor{white} & Ham.-Weiner & & & & & & & & \\
        % \multirow{-5}{*}{\cellcolor{white} \rotatebox[origin=c]{90}{\textbf{Robot B}}}
        % & NLARX & & & & & & & & \\
        % \hline
    \end{tabular}
    \label{tab:RMSE}
\end{table}
\section{Discussion and Conclusion}
\label{sec:conclusion}

%% Summary of what we've shown
We have successfully applied a system identification technique based on Koopman operator theory to a soft robot and shown that the model generated outperforms those constructed by several other state-of-the-art nonlinear system identification methods.
% This data-driven approach enables the identification of dynamic models of soft robots which can be used in control.
Perhaps unsurprisingly, the linear state space model was unable to capture the nonlinear dynamics of the robot as well as the Koopman model.
%% Why does Koopman model do better than nonlinear models
As for the nonlinear models, there are several likely reasons why the performance of the Koopman model was superior.
Since the Koopman model is a state space model, simulations can be initialized from the same initial condition as the real system.
This is not the case for the other learned nonlinear models which do not have an internal state corresponding to the physical state of the robot.
Rather, they act as black-box models only capable of mapping inputs to outputs.
% serve as black-box models only capable of mapping inputs to outputs without an internal state that corresponds to the physical state of the robot.
% For simulations over short time horizons, this effect can contribute significantly to error.

%% Other models rely on manual tuning so success is highly dependent on training parameters chosen
Another advantage of the Koopman model is that its quality does not depend on an initial model estimate or tuning parameters.
By iterating over the set of all initializations and tuning parameters, one may be able to generate better performing models than those shown; unfortunately, this multivariate trial-and-error process may not affect results in a predictable way.
In contrast, the only tuning parameter involved in the Koopman method is the maximum degree of the monomial basis functions, which has a magnitude that is directly proportional to model accuracy. 

%% Challenges and how it could be improved
While the results here are promising, there are practical challenges to extending the Koopman approach to higher dimensional systems.
As the dimension of the state space increases, so does the size of the monomial basis set of the finite-dimensional subspace of observables.
This greatly increases the size of the matrix equations that must be solved, leading to computational intractability for sufficiently high dimensional systems.
However, if some information about the system is known beforehand, this issue could be counteracted by choosing a more suitable basis for the observables.
For example, if the system exhibits oscillatory motion, a lower dimensional fourier basis may be more suitable than a monomial basis to represent the behavior.
Such an extension of the method is left to future work.

%% Numerical challenges
% There are still numerical challenges to applying this approach to arbitrary systems.
% As stated previously, the method fails if not enough data points are supplied, but if the training data set is too large, it becomes impractical to implement due to computer memory constraints.
% Therefore, a focus of future work will be to investigate how to generate a representative sampling of system behavior of sufficient size to ensure the method can be successfully implemented.
% This approach also suffers from the curse of dimensionality...

%% Future work / remaining challenges
Soft robots are notoriously difficult to model, but amenable to large-scale data collection and data-driven modeling methods. 
This paper demonstrates the utility of Koopman operator theory to make accurate nonlinear dynamical models easier to construct, enabling  the  rapid  development  of control strategies that exploit the unique characteristics of soft robots.
Future work will aim to generalize this approach to higher dimensional models, non-polynomial models, and models that account for external loading and contact forces.

%                                   % the textheight of the last page by a suitable amount.
%                                   % This command does not take effect until the next page
%                                   % so it should come on the page before the last. Make
%                                   % sure that you do not shorten the textheight too much.

% \input{sections/acknowledgement.tex}

\bibliographystyle{IEEEtran}
\bibliography{references}

\begin{thebibliography}{10}
\providecommand{\url}[1]{#1}
\csname url@rmstyle\endcsname
\providecommand{\newblock}{\relax}
\providecommand{\bibinfo}[2]{#2}
\providecommand\BIBentrySTDinterwordspacing{\spaceskip=0pt\relax}
\providecommand\BIBentryALTinterwordstretchfactor{4}
\providecommand\BIBentryALTinterwordspacing{\spaceskip=\fontdimen2\font plus
\BIBentryALTinterwordstretchfactor\fontdimen3\font minus
  \fontdimen4\font\relax}
\providecommand\BIBforeignlanguage[2]{{%
\expandafter\ifx\csname l@#1\endcsname\relax
\typeout{** WARNING: IEEEtran.bst: No hyphenation pattern has been}%
\typeout{** loaded for the language `#1'. Using the pattern for}%
\typeout{** the default language instead.}%
\else
\language=\csname l@#1\endcsname
\fi
#2}}

\bibitem{rus2015design}
D.~Rus and M.~T. Tolley, ``Design, fabrication and control of soft robots,''
  \emph{Nature}, vol. 521, no. 7553, p. 467, 2015.

\bibitem{majidi2014soft}
C.~Majidi, ``Soft robotics: a perspective—current trends and prospects for
  the future,'' \emph{Soft Robotics}, vol.~1, no.~1, pp. 5--11, 2014.

\bibitem{lipson2014challenges}
H.~Lipson, ``Challenges and opportunities for design, simulation, and
  fabrication of soft robots,'' \emph{Soft Robotics}, vol.~1, no.~1, pp.
  21--27, 2014.

\bibitem{felt2018modeling}
W.~Felt, S.~Lu, and C.~D. Remy, ``Modeling and design of “smart braid”
  inductance sensors for fiber-reinforced elastomeric enclosures,'' \emph{IEEE
  Sensors Journal}, vol.~18, no.~7, pp. 2827--2835, 2018.

\bibitem{felt2017inductance}
W.~Felt, M.~J. Telleria, T.~F. Allen, G.~Hein, J.~B. Pompa, K.~Albert, and
  C.~D. Remy, ``An inductance-based sensing system for bellows-driven continuum
  joints in soft robots,'' \emph{Autonomous Robots}, pp. 1--14, 2017.

\bibitem{yang2013gauge}
S.~Yang and N.~Lu, ``Gauge factor and stretchability of silicon-on-polymer
  strain gauges,'' \emph{Sensors}, vol.~13, no.~7, pp. 8577--8594, 2013.

\bibitem{kim2011epidermal}
D.-H. Kim, N.~Lu, R.~Ma, Y.-S. Kim, R.-H. Kim, S.~Wang, J.~Wu, S.~M. Won,
  H.~Tao, A.~Islam, \emph{et~al.}, ``Epidermal electronics,'' \emph{science},
  vol. 333, no. 6044, pp. 838--843, 2011.

\bibitem{della2017controlling}
C.~Della~Santina, M.~Bianchi, G.~Grioli, F.~Angelini, M.~Catalano, M.~Garabini,
  and A.~Bicchi, ``Controlling soft robots: balancing feedback and feedforward
  elements,'' \emph{IEEE Robotics \& Automation Magazine}, vol.~24, no.~3, pp.
  75--83, 2017.

\bibitem{bishop2015design}
J.~Bishop-Moser and S.~Kota, ``Design and modeling of generalized
  fiber-reinforced pneumatic soft actuators,'' \emph{IEEE Transactions on
  Robotics}, vol.~31, no.~3, pp. 536--545, 2015.

\bibitem{bruder2018iros}
D.~Bruder, A.~Sedal, R.~Vasudevan, and C.~D. Remy, ``Force generation by
  parallel combinations of fiber-reinforced fluid-driven actuators,''
  \emph{IEEE Robotics and Automation Letters}, vol.~3, no.~4, pp. 3999--4006,
  Oct 2018.

\bibitem{renda2014dynamic}
F.~Renda, M.~Giorelli, M.~Calisti, M.~Cianchetti, and C.~Laschi, ``Dynamic
  model of a multibending soft robot arm driven by cables,'' \emph{IEEE
  Transactions on Robotics}, vol.~30, no.~5, pp. 1109--1122, 2014.

\bibitem{felt2018closed}
W.~Felt and C.~D. Remy, ``A closed-form kinematic model for fiber-reinforced
  elastomeric enclosures,'' \emph{Journal of Mechanisms and Robotics}, vol.~10,
  no.~1, p. 014501, 2018.

\bibitem{neppalli2009closed}
S.~Neppalli, M.~A. Csencsits, B.~A. Jones, and I.~D. Walker, ``Closed-form
  inverse kinematics for continuum manipulators,'' \emph{Advanced Robotics},
  vol.~23, no.~15, pp. 2077--2091, 2009.

\bibitem{webster2010design}
R.~J. Webster~III and B.~A. Jones, ``Design and kinematic modeling of constant
  curvature continuum robots: A review,'' \emph{The International Journal of
  Robotics Research}, vol.~29, no.~13, pp. 1661--1683, 2010.

\bibitem{jones2006kinematics}
B.~A. Jones and I.~D. Walker, ``Kinematics for multisection continuum robots,''
  \emph{IEEE Transactions on Robotics}, vol.~22, no.~1, pp. 43--55, 2006.

\bibitem{george2018control}
T.~George~Thuruthel, Y.~Ansari, E.~Falotico, and C.~Laschi, ``Control
  strategies for soft robotic manipulators: A survey,'' \emph{Soft robotics},
  vol.~5, no.~2, pp. 149--163, 2018.

\bibitem{gravagne2003large}
I.~A. Gravagne, C.~D. Rahn, and I.~D. Walker, ``Large deflection dynamics and
  control for planar continuum robots,'' \emph{IEEE/ASME transactions on
  mechatronics}, vol.~8, no.~2, pp. 299--307, 2003.

\bibitem{trivedi2008geometrically}
D.~Trivedi, A.~Lotfi, and C.~D. Rahn, ``Geometrically exact models for soft
  robotic manipulators,'' \emph{IEEE Transactions on Robotics}, vol.~24, no.~4,
  pp. 773--780, 2008.

\bibitem{bruder2017model}
D.~Bruder, A.~Sedal, J.~Bishop-Moser, S.~Kota, and R.~Vasudevan, ``Model based
  control of fiber reinforced elastofluidic enclosures,'' in \emph{Robotics and
  Automation (ICRA), 2017 IEEE International Conference on}.\hskip 1em plus
  0.5em minus 0.4em\relax IEEE, 2017, pp. 5539--5544.

\bibitem{sedal2017constitutive}
A.~Sedal, D.~Bruder, J.~Bishop-Moser, R.~Vasudevan, and S.~Kota, ``A
  constitutive model for torsional loads on fluid-driven soft robots,'' in
  \emph{ASME 2017 International Design Engineering Technical Conferences and
  Computers and Information in Engineering Conference}.\hskip 1em plus 0.5em
  minus 0.4em\relax American Society of Mechanical Engineers, 2017, pp.
  V05AT08A016--V05AT08A016.

\bibitem{bishop2012parallel}
J.~Bishop-Moser, G.~Krishnan, C.~Kim, and S.~Kota, ``Design of soft robotic
  actuators using fluid-filled fiber-reinforced elastomeric enclosures in
  parallel combinations,'' in \emph{Intelligent Robots and Systems (IROS), 2012
  IEEE/RSJ International Conference on}.\hskip 1em plus 0.5em minus 0.4em\relax
  IEEE, 2012, pp. 4264--4269.

\bibitem{boyd2004convex}
S.~Boyd and L.~Vandenberghe, \emph{Convex optimization}.\hskip 1em plus 0.5em
  minus 0.4em\relax Cambridge university press, 2004.

\bibitem{gillespie2018learning}
M.~T. Gillespie, C.~M. Best, E.~C. Townsend, D.~Wingate, and M.~D. Killpack,
  ``Learning nonlinear dynamic models of soft robots for model predictive
  control with neural networks,'' in \emph{2018 IEEE International Conference
  on Soft Robotics (RoboSoft)}.\hskip 1em plus 0.5em minus 0.4em\relax IEEE,
  2018.

\bibitem{ljung1987system}
L.~Ljung, \emph{System identification: theory for the user}.\hskip 1em plus
  0.5em minus 0.4em\relax Prentice-hall, 1987.

\bibitem{budivsic2012applied}
M.~Budi{\v{s}}i{\'c}, R.~Mohr, and I.~Mezi{\'c}, ``Applied koopmanism,''
  \emph{Chaos: An Interdisciplinary Journal of Nonlinear Science}, vol.~22,
  no.~4, p. 047510, 2012.

\bibitem{williams2015data}
M.~O. Williams, I.~G. Kevrekidis, and C.~W. Rowley, ``A data--driven
  approximation of the koopman operator: Extending dynamic mode
  decomposition,'' \emph{Journal of Nonlinear Science}, vol.~25, no.~6, pp.
  1307--1346, 2015.

\bibitem{mauroy2016linear}
A.~Mauroy and J.~Goncalves, ``Linear identification of nonlinear systems: A
  lifting technique based on the koopman operator,'' \emph{arXiv preprint
  arXiv:1605.04457}, 2016.

\bibitem{mauroy2017koopman}
------, ``Koopman-based lifting techniques for nonlinear systems
  identification,'' \emph{arXiv preprint arXiv:1709.02003}, 2017.

\bibitem{Abraham-RSS-17}
I.~Abraham, G.~de~la Torre, and T.~Murphey, ``Model-based control using koopman
  operators,'' in \emph{Proceedings of Robotics: Science and Systems},
  Cambridge, Massachusetts, July 2017.

\bibitem{lasota2013chaos}
A.~Lasota and M.~C. Mackey, \emph{Chaos, fractals, and noise: stochastic
  aspects of dynamics}.\hskip 1em plus 0.5em minus 0.4em\relax Springer Science
  \& Business Media, 2013, vol.~97.

\bibitem{higham2008functions}
N.~J. Higham, \emph{Functions of matrices: theory and computation}.\hskip 1em
  plus 0.5em minus 0.4em\relax Siam, 2008, vol. 104.

\bibitem{MATLAB:2017}
MATLAB, \emph{version 7.10.0 (R2017a)}.\hskip 1em plus 0.5em minus 0.4em\relax
  Natick, Massachusetts: The MathWorks Inc., 2017.

\end{thebibliography}

\end{document}